\documentclass[letterpaper, 10 pt, journal, twoside]{ieeetran}

\IEEEoverridecommandlockouts                              

\usepackage{microtype}
\usepackage{graphicx}
\usepackage{subcaption}
\usepackage{booktabs} 
\usepackage{amssymb}
\usepackage{amsmath}
\usepackage{multirow}
\usepackage{url}
\usepackage{hyperref}

\usepackage{enumitem}
\usepackage{balance}
\usepackage{threeparttable}

\newcommand{\eg}{\textit{e.g.}\xspace}
\newcommand{\ie}{\textit{i.e.}\xspace}

\usepackage{makecell}
\usepackage[table]{xcolor}
\definecolor{bestcolor}{gray}{.9}
\newcommand{\bestcell}[1]{\cellcolor{bestcolor}{#1}}

\newcommand{\rebuttal}[1]{\textcolor{black}{#1}}

\usepackage{xspace}
\newcommand{\alias}{COMPACT-VA\xspace}

\title{
Planning-aligned Token Compression for Long-Context\\Autonomous Driving
}

\author{Zhixuan Liang$^{1,2*}$, Yuxiao Chen$^{1}$, Yurong You$^{1}$, Peter Karkus$^{1}$, Wenhao Ding$^{1}$,\\
Boyi Li$^{1}$, Alexander Popov$^{1}$, Yan Wang$^{1}$, Maximilian Igl$^{1}$, Yiming Li$^{1}$,\\
Danfei Xu$^{1,4}$, Nikolai Smolyanskiy$^{1}$, Boris Ivanovic$^{1}$, Ping Luo$^{2\dagger}$, Marco Pavone$^{1,3\dagger}$%
\thanks{Manuscript received: March 16, 2026; Revised June 23, 2026; Accepted August 9, 2026.
This paper was recommended for publication by Editor Joerg Stueckler upon evaluation of the Associate Editor and Reviewers'~comments.
}
\thanks{$^{1}$Z.~Liang, Y.~Chen, Y.~You, P.~Karkus, W.~Ding, B.~Li, 
        A.~Popov, Y.~Wang, M.~Igl, Y.~Li, D.~Xu, N.~Smolyanskiy, 
        B.~Ivanovic, and M.~Pavone are with NVIDIA 
        {\tt\footnotesize \{yuxiaoc, bivanovic, mpavone\}@nvidia.com}}%
\thanks{$^{2}$Z.~Liang and P.~Luo are with School of Computing and Data Science, The University of Hong Kong {\tt\footnotesize \{zxliang, pluo\}@cs.hku.hk}}%
\thanks{$^{3}$M.~Pavone is also affiliated with Stanford University}%
\thanks{$^{4}$D.~Xu is also affiliated with Georgia Institute of Technology}%
\thanks{*This work was done during Zhixuan's internship at NVIDIA Research.}%
\thanks{$^{\dagger}$Ping Luo and Marco Pavone are corresponding authors.}
\thanks{Digital Object Identifier (DOI): see top of this page.}
}

\begin{document}

\maketitle

\markboth{IEEE Robotics and Automation Letters. Preprint Version. Accepted August, 2026}
{Liang \MakeLowercase{\textit{et al.}}: Planning-aligned Token Compression for Long-Context Autonomous Driving} 

\begin{abstract}

Monolithic vision-action models represent an emerging paradigm in autonomous driving. However, this architecture produces token sequences that quickly exceed real-time computational budgets when encoding extended temporal context for complex interactions.
While approaches like linear transformers and external memory try to make the context lightweight, token compression is most compatible with the architecture as it requires no backbone modifications. 
%
Yet existing compression adopts rule-based heuristics like temporal decay, decoupled from planning, risking loss of decision-critical information.
%
We propose \alias, a planning-aligned working memory framework built on conditional VQ-VAE, compressing extended context into bounded representations. Compression is conditioned on both historical trajectory and a learned planning intent that the posterior encoder distills from future trajectories during training, while the prior encoder learns to predict it from compressed observations.
The compressed memory, concatenated with the predicted latent, feeds the policy for end-to-end optimization, planning with retained decision-critical information.
We evaluate on high-signal dynamic scenarios where historical context is most critical for behavior (\eg, stop, yield, or proceed) correctness, and accordingly design behavioral metrics.
Under comparable token budgets, we achieve~$>$6\%~improvement~(68.3\%)~on success with consistent gains across metrics.
Ablations validate the effectiveness of planning-aligned coupling. Closed-loop evaluation confirms \alias maintain general driving performance with 3.3$\times$ speedup and 2.7$\times$ memory reduction over uncompressed baseline.

\end{abstract}


\section{INTRODUCTION}

\IEEEPARstart{V}{ision}-action (VA) and vision-language-action (VLA) policies~\cite{wang2025alpamayo} represent the latest paradigm in autonomous driving, directly mapping all modality inputs to vehicle trajectories through a unified transformer backbone. Unlike modular pipelines with separate perception–prediction–planning stages~\cite{hu2023uniad,weng2024para} that maintain explicit state representations (\eg, bounding boxes) across modules, VLA policies instead encode the entire history directly within the observation token sequence, enabling fully end-to-end learning.

\begin{figure}[tb]
  \begin{center}
    \centerline{\includegraphics[width=0.98\linewidth]{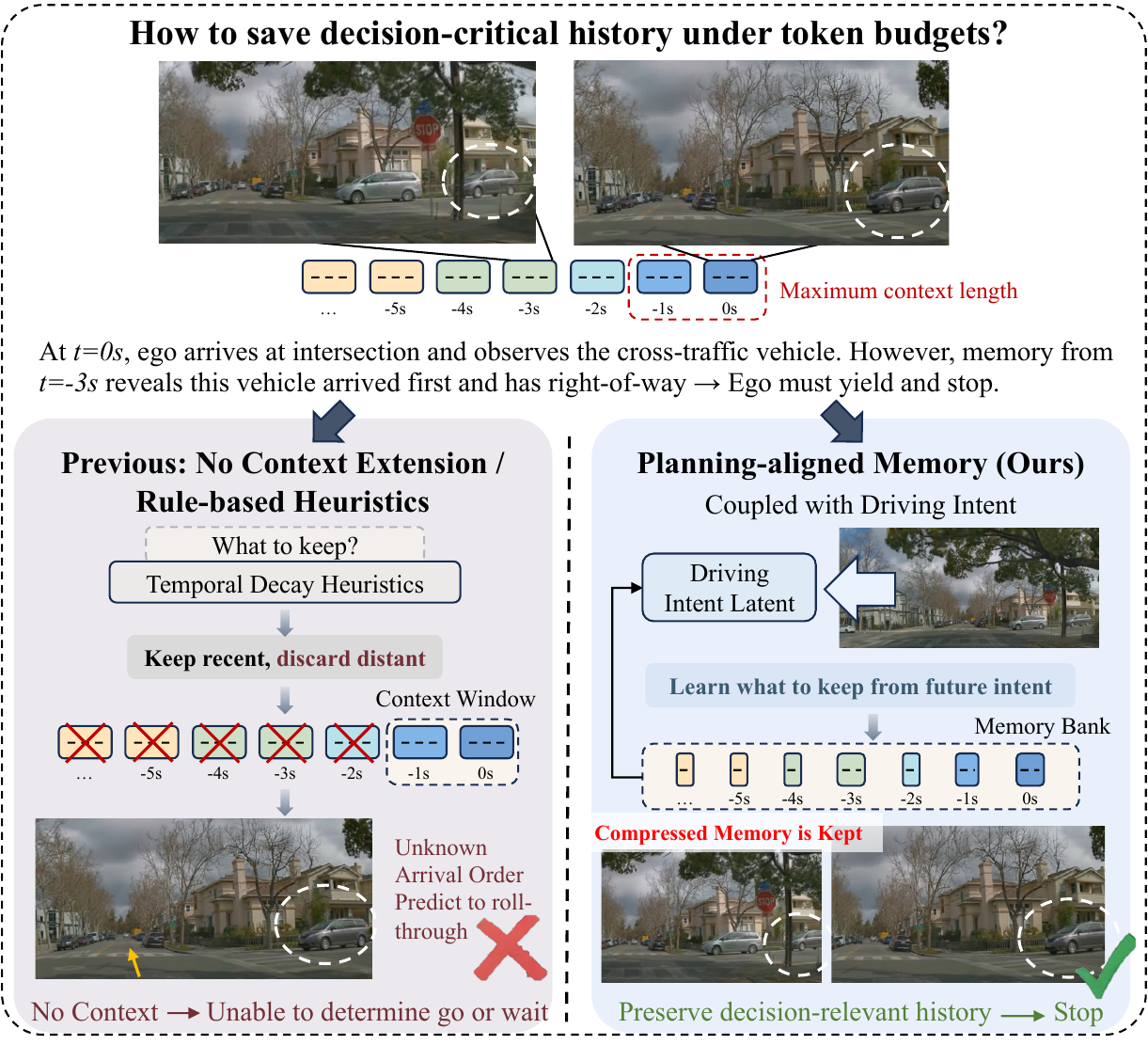}}
    \vspace{-2pt}
    \caption{\textbf{From limited context to planning-aligned memory.} At an all-way sign intersection, determining right-of-way requires observing traffic over extended periods. Rule-based temporal decay with a default 2s window discards distant observations, losing information about which vehicle arrived first (happended at t=-3s) and causing incorrect go decisions (62\% success). Our planning-aligned compression retains decision-critical cues through coupling with driving intent, preserving arrival order information for correct behavior (68.3\% success).}
    
    \label{fig:teaser}
  \end{center}
  \vspace{-20pt}
\end{figure}

Scaling these models to handle complex driving scenarios introduces a fundamental challenge that a longer history quickly makes the visual sequence length exceed the real-time computational budgets. Several approaches have been proposed to address this, including linear transformers~\cite{gu2024mamba} and external memory modules~\cite{shi2025memoryvla}. Among them, token compression has emerged as the most practical solution, as it requires no backbone modifications and naturally supports the medium term memory horizon (on the order of tens of seconds) that driving decisions demand~\cite{al2024autonomous}. Despite it, existing compression methods rely on rule-based heuristics such as temporal decay, retaining recent frames while discarding older ones. These strategies are decoupled from the planning objective and cannot distinguish decision-critical history from other redundant information.
As illustrated in Fig.~\ref{fig:teaser}, rule-based compression with temporal decay discards historical information about arrival order, while our planning-aligned approach learns to retain it by coupling compression with driving intent prediction.


On this basis, we propose COMpression via Planning-Aligned Context Token framework (\textit{abbr.}~\alias), a working memory mechanism that learns what history to retain through closed-loop optimization of driving performance. 
We instantiate this framework with a Q-former-based learnable compression module~\cite{li2023blip2} with a conditional variational auto-encoder (cVAE) with vector quantization (VQ), introducing a compact discrete intent representation bridging compressed observations and trajectory prediction. 
During compression, learnable query tokens are interleaved with raw observation tokens, history trajectory, and learned latent through self-attention. This design enables compression conditioned on both past trajectory context and future driving intent, constituting \emph{planning-aligned working memory}. Compressing budgets are organized in a hierarchical FIFO buffer retaining more tokens for recent frames and fewer for distant ones.
%
%
The compressed memory then feeds the policy backbone for trajectory prediction (detailed in Sec.~\ref{sec:method}). Through joint optimization, the model learns decision-relevant memory without hand-crafted rules.
    
To evaluate our approach, we focus on high-signal scenarios where extended historical context plays a prominent role, such as all-way stops, dynamic occlusion and unprotected turns. These scenarios share a fundamental characteristic that good behaviors depend on making right discrete decisions (\ie, whether the vehicle stops when required and proceeds when appropriate) rather than trajectory optimization over a continuous spectrum. We design behavioral metrics including \textit{stop/go success rates, roll-through rates, and stop position/duration errors} tailored to these decision-critical scenarios.

Under comparable token budgets, \alias achieves 68.3\% go success rate versus $\sim$62.0\% for baselines (+6.3\%) and reduces safety-critical roll-throughs by 22\%. Compared to compression without planning alignment (65.6\%), \alias demonstrates +2.7\% improvement, confirming that coupling compression with planning forces the model to retrain decision-critical information. Ablations validate that each of the sub-components, such as hierarchical compression, history conditioning, and planning coupling, contribute to improvements in high-signal scenarios. Closed-loop evaluation on 910 diverse scenarios confirms \alias maintains general driving competence while achieving 3.3$\times$ speedup and 2.7$\times$ memory reduction over uncompressed scheme.

In summary, our contributions are threefold:
(1) We identify high-signal dynamic scenarios, including all-way stops, dynamic occlusion and unprotected turns as critical testbeds for memory-dependent driving, and introduce behavioral metrics assessing decision correctness beyond trajectory displacement.
(2) We propose a planning-aligned working memory for VLA driving policies built on a conditional VAE, where compression quality is explicitly tied to trajectory prediction through a variational objective, enabling end-to-end learning of task-relevant compression.
(3) Experiments demonstrate sizable improvements under comparable token budgets, and achieve consistent gains across all metrics. Ablations validate each component's contribution to performance.

\section{RELATED WORK}

\subsection{Long-term Memory for Physical AI}

Long-term memory is essential for physical AI systems under partial observability, including manipulation, navigation, and autonomous driving. Classical approaches maintain world state through explicit estimation (\eg, SLAM) or belief-state updates~\cite{thrun2002probabilistic}, while latent world models encode history for long-horizon reasoning~\cite{hafner2019dreamer,kaelbling1998pomdp}. Recent VLA policies revisit external memory modules for manipulation~\cite{shi2025memoryvla,driess2023palme}, and driving-specific VLAs adopt similar designs under open-data training regimes~\cite{chi2026impromptu}.
%
%
In contrast, autonomous driving and other high-rate embodied systems require memory mechanisms that are both computationally bounded and causally grounded, that means the mechanism should retain only task-relevant history while discards redundant observations. Our approach aligns with this view of memory as a compressed, persistent interface between past and present observations~\cite{rae2019compressive}.

\vspace{-7pt}
\subsection{Long-Context Transformers with Recurrence}

A prominent direction for long-context modeling is to augment attention with explicit recurrence or external memory. Transformer-XL introduces segment-level recurrence~\cite{dai2019transformerxl}, and the Compressive Transformer maintains dual memories to compress old activations to retain salient information~\cite{rae2019compressive}. 
A separate but related line explores memory as a learnable component updated at inference. TTT layers~\cite{sun2024ttt} make hidden states an expressive model updated via self-supervised learning, while Titans memorize during inference to complement local attention~\cite{behrouz2025titans}. These works reinforce long-term memory to be an explicit, persistent bank rather than an ever-growing context window. 
Our compression draws inspiration from this compressive-memory view and adapts it to driving policy with multimodal inputs. Moreover, we introduce a variational objective tying compression directly to planning performance, ensuring decision-relevant information retained.
%
%

\vspace{-7pt}
\subsection{Token Compression and Efficient Seq Architectures}

Orthogonal to recurrence, token-level compression reduces computational cost by merging or pruning tokens. ToMe~\cite{bolya2023tome} merges similar tokens based on feature similarity, while StreamingLLM~\cite{xiao2024streamingllm} retains initial tokens as attention sinks to stabilize long-sequence decoding. Sparse attention, pruning, and pooling strategies have been widely explored in vision and video transformers. To address quadratic attention cost, sparse-attention transformers (\eg, Longformer~\cite{beltagy2020longformer}, BigBird~\cite{zaheer2020bigbird}) limit patterns to local windows plus global tokens, while state-space models (\eg, Mamba~\cite{gu2024mamba}) achieve linear scaling through selective mechanisms. Hybrid designs (\eg, Hymba~\cite{dong2025hymba}) combine attention with SSM heads for efficient context summarization. 
While effective at reducing cost, these methods compress tokens agnostic to planning objectives, potentially discarding task-critical information. Inspired by Hint-AD~\cite{ding2025hint}, which shows that planning-aligned representations improve both interpretability and performance in end-to-end driving, our method couples compression directly with the planning objective to retain only planning-relevant historical context.


\begin{figure*}[tb]
  \begin{center}
    \centerline{\includegraphics[width=0.99\linewidth]{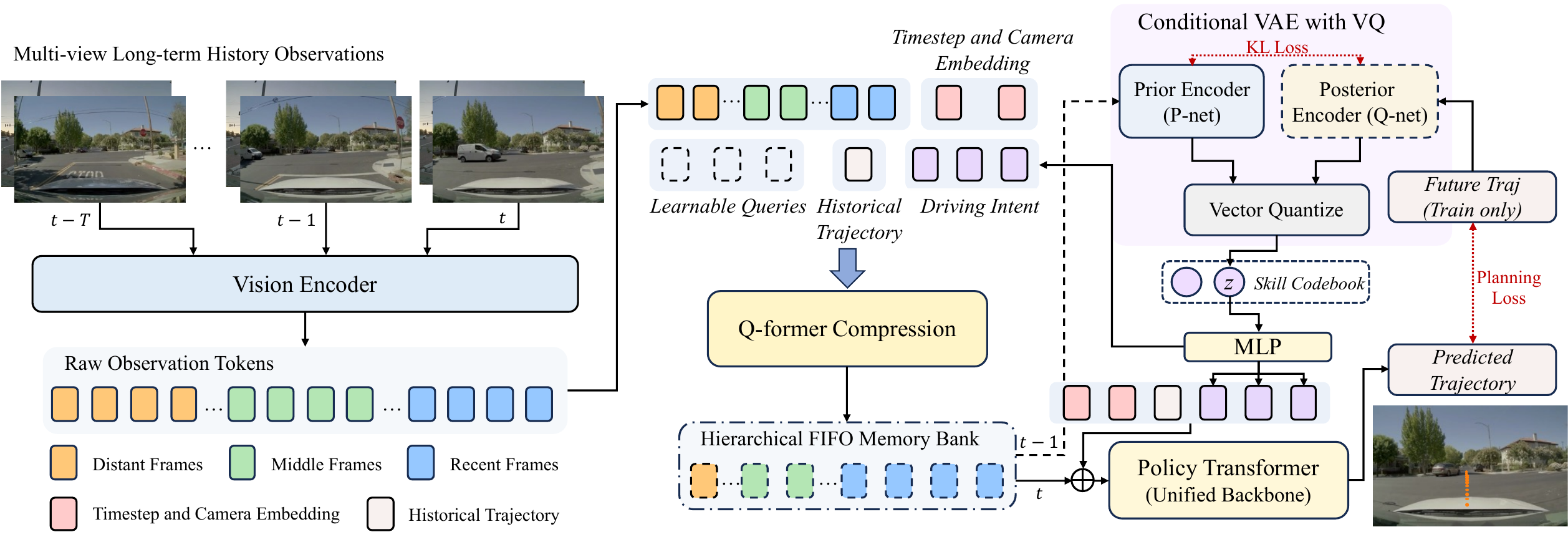}}
    \caption{\textbf{Overall architecture of \alias.} Multi-view raw observation tokens from current and past are compressed via Q-former conditioning on camera embeddings, historical trajectory and a learned driving intent latent. The posterior encoder (Q-net) distills intent from future trajectories during training, while the prior encoder (P-net) predicts this latent from compressed observations alone. The hierarchical FIFO memory bank, concatenated with the predicted latent as a special token, feeds the policy transformer, with end-to-end optimization coupling compression quality to planning performance.}
    \label{fig:overall}
  \end{center}
  \vspace{-22pt}
\end{figure*}

\section{METHOD}
\label{sec:method}

\subsection{Unified Vision-Action Model Backbone}
\label{sec:va_backbone}

Our approach builds upon the unified vision-action (VA) policy variant of Alpamayo~\cite{wang2025alpamayo}, which consists of three core components: a vision encoder processing multi-camera observations into visual tokens, a transformer backbone performing temporal reasoning without text instruction, and a trajectory decoder generating future vehicle motion.

Multi-camera images from the current and past $T$ timestamps are encoded into visual tokens through a pre-trained vision encoder (\eg, DINOv2~\cite{dinov2}). Each image produces $N_{\text{img}}$ tokens. With $N_{\text{cam}}$ cameras per timestep for multi-view observation, the raw vision token count grows as:
\vspace{-2pt}
\begin{equation}
N_{\text{raw}} = T \times N_{\text{cam}} \times N_{\text{img}}.
\vspace{-2pt}
\end{equation}
These vision tokens, along with temporal positional embeddings and camera-specific embeddings, are concatenated with encoded historical trajectory information, and then fed into a causal transformer backbone. For historical trajectories, we apply sinusoidal positional embeddings with MLP compression to produce a single continuous token representing the ego history.
%
For future trajectory, we adopt an FSQ-based tokenizer~\cite{FSQ}, compressing future waypoints into discrete tokens via finite scalar quantization, enabling autoregressive generation while maintaining reconstruction quality (\textit{c.f.} Alpamayo~\cite{wang2025alpamayo}).


While this unified architecture eliminates explicit intermediate modules, the sequence length grows linearly with the context length. For complex driving scenarios requiring extended temporal context, the token count grows substantially, easily surpassing typical VLM context windows. Without effective compression, the quadratic transformer attention cost $\mathcal{O}(N_{\text{raw}}^2)$ becomes intractable for deployment.

\vspace{-7pt}
\subsection{Learned Hierarchical Temporal Context Buffering}
\label{sec:hierarchical}

To manage token sequence growth while preserving temporal information, we compress raw observations through learned query-based aggregation organized into a hierarchical memory bank. The overall architecture of \alias is shown in Fig.~\ref{fig:overall}. We first describe the hierarchical compression module, then introduce the planning-aligned variational coupling in Sec.~\ref{sec:vae_compression}, which determines what information to retain.


\vspace{2pt}\textbf{Hierarchical Buffer Structure.}
The observation history spanning $T$ timesteps is organized into $K$ compression layers $\{L_1, L_2, \ldots, L_K\}$, each applying different compression ratios to balance token efficiency and information preservation. Each layer $L_k$ contains $n_k$ consecutive frames ($\sum_{k=1}^{K} n_k = T$). Compression is applied hierarchically in a cascading manner where frames are first encoded into $N_{\text{img}}$ tokens per camera, then progressively compressed through the layers, with layer $L_k$ producing $\lfloor N_{\text{img}} / r_k \rfloor$ tokens per camera per frame, where $r_k$ is the cumulative compression ratio (\ie, the product of compression factors from layer 1 to layer $k$ relative to the original $N_{\text{img}}$). The total compressed token count becomes:
\vspace{-7pt}
\begin{equation}
N_{\text{compressed}} = \sum_{k=1}^{K} n_k \cdot N_{\text{cam}} \cdot \lfloor N_{\text{img}} / r_k \rfloor.
\vspace{-5pt}
\end{equation}

We employ a multi-layer hierarchy following a temporal-decay heuristic where the most recent layer retains full token resolution (no compression), the intermediate layer applies moderate compression, and the distant layer uses aggressive compression. The layer durations are proportioned such that recent history occupies a smaller temporal window but retains higher token density, while distant history spans a longer period with sparse representation. This design achieves substantial compression while preserving fine-grained information where behavioral cues are most critical. Specific configuration details are provided in Sec.~\ref{sec:impl_details}.

\vspace{2pt}\textbf{Learned Compression via Q-former.}
The compression at each level is realized through a Q-former module that takes the buffered observation tokens $\{o_t^{\text{buffer}}\}$ as input, along with their associated timestep embeddings $e_{\text{time}}$ and camera embeddings $e_{\text{cam}}$. For frames at level $L_k$, we concatenate $\lfloor N_{\text{img}} / r_k \rfloor$ learnable query tokens with the raw observation tokens and other embeddings, then process this combined sequence through self-attention in an MMDiT manner~\cite{stable_diffusion_3}. The query tokens selectively aggregate task-relevant visual features through bidirectional attention, producing the compressed representation for each frame.

After compression, the tokens from all levels are reordered chronologically, from early to late frames. To accommodate the hierarchical compression structure, we employ the RoPE positional embeddings~\cite{rope} from Alpamayo such that, for tokens at level $L_k$ with compression ratio $r_k$, the RoPE frequency step is scaled by $r_k$. This keeps positional encodings consistent across compression levels and aligned with the uncompressed sequence.

Rather than relying on hand-crafted rules, the compression is learned end-to-end. While the hierarchical buffer structure provides an inductive bias for temporal decay, the query tokens adaptively determine which visual features to retain. The resulting compressed tokens form the \emph{trajectory-conditioned memory}. Next, we introduce the variational framework that enables planning-aligned compression.

\subsection{Planning-aligned Variational Token Compression}
\label{sec:vae_compression}

While the hierarchical compression (Sec.~\ref{sec:hierarchical}) reduces token count, it does not explicitly couple compression with the planning objective to determine what information to retain based on downstream planning needs. We introduce a conditional variational auto-encoder (cVAE) framework with vector quantization (VQ) that addresses this by coupling compression quality with trajectory prediction. The key idea is to distill driving intent from future trajectories into a compact discrete latent $z \in \mathbb{R}^d$, then train compressed observations to be sufficient for predicting this latent, ensuring retention of decision-critical historical cues (Fig.~\ref{fig:overall}).

\vspace{2pt}\textbf{Variational Encoder Architecture.}
We employ two encoders with distinct roles during training and inference.
Both encoders use lightweight architectures to remain computationally efficient relative to the policy backbone.


\textit{1) Posterior encoder $q_\phi(z \mid o, \tau_{\text{future}})$ (training only):} Extracts driving intent from future trajectories. Following Alpamayo~\cite{wang2025alpamayo}, ground-truth trajectories are first converted to unicycle control sequences $(a, \kappa)$ for acceleration and curvature via least-squares optimization with Tikhonov regularization~\cite{golub1999tikhonov} to attenuate high-frequency noise, then uniformly quantized into discrete tokens. We find that this approach better captures trajectory dynamics compared to FSQ tokenization. These tokens are then compressed via an MLP into $N_{\text{agg}}$ tokens including one global token and $N_{\text{local}}$ local tokens from uniformly divided segments. A small transformer outputs mean $\mu_\phi$ and log-variance $\log \sigma^2_\phi$ parameterizing a Gaussian distribution $\mathcal{N}(\mu_\phi, \sigma^2_\phi)$, from which latent $z_q$ is sampled.

\textit{2) Prior encoder $p_\theta(z \mid o_{\text{compressed}})$ (training and inference):} This encoder predicts the driving intent latent using only the compressed observations $o_{\text{compressed}}$ from the Q-former (Sec.~\ref{sec:hierarchical}), without access to future information. It processes the compressed tokens through attention pooling followed by MLPs to produce $z_p$. During training, it learns to match the posterior distribution; at inference, it operates independently to predict driving intent from historical context alone.

\textit{3) Vector Quantization:} Both $z_q$ and $z_p$ are mapped to a shared discrete codebook via $\text{argmin}_k \|z - c_k\|$, yielding quantized embedding $z_{\text{skill}} = c_{i^*}$. Following~\cite{van2017neural}, gradient flow uses the straight-through estimator $z_{\text{skill}} = z + (c_{i^*} - z).\text{detach}()$, with commitment loss $\mathcal{L}_{\text{commit}} = \|z - c_{i^*}\|^2$ encouraging alignment with the codebook.


\vspace{2pt}\textbf{Policy Input Composition.}
The discrete skill embedding $z_{\text{skill}}$ obtained from VQ, using prior encoder's $z_p$ at training and inference, is re-projected through a learned linear layer and appended as a special token. This token is concatenated with the trajectory-conditioned memory $o_{\text{compressed}}$, the historical trajectory token (Sec.~\ref{sec:va_backbone}), and re-applied timestep and camera embeddings. This combined sequence is fed into the unified transformer backbone following Alpamayo~\cite{wang2025alpamayo}, which autoregressively predicts the future trajectory tokens.

\vspace{2pt}\textbf{End-to-End Training.}
The entire system consisting of Q-former compression, prior/posterior encoders, VQ codebook, and policy backbone, is optimized end-to-end with,  
\begin{align}
    \mathcal{L} = \mathcal{L}_{\text{traj}} &+ \lambda_{\text{KL}} \cdot D_{\text{KL}}\Big(q_\phi(z | o, \tau_{\text{future}}) \,\|\, p_\theta(z | o_{\text{compressed}})\Big) \nonumber \\ 
    &+ \lambda_{\text{commit}} \cdot \mathcal{L}_{\text{commit}},\vspace{-13pt}
\end{align}  
where $\mathcal{L}_{\text{traj}}$ is the cross-entropy loss over discrete future trajectory tokens, and the KL divergence encourages the prior distribution to match the posterior.

During training, the policy conditions on the latent sampled from the \emph{prior} encoder $z_p$, instead of the posterior $z_q$, ensuring consistency between training and inference. This design creates a closed-loop coupling between compression and planning. If the Q-former discards any decision-critical historical information, the prior encoder cannot accurately predict the latent inferred from future trajectories, resulting in both high KL divergence and degraded trajectory prediction. Through this joint optimization, the model implicitly discovers what historical information matters for downstream decisions, without requiring hand-crafted retention rules.

\vspace{2pt}\textbf{Inference.}
At test time, only the prior pathway is active. The model compresses observations via Q-former, predicts the latent $z_p$ from trajectory-conditioned memory, quantizes it via VQ to retrieve the discrete skill embedding, re-projects and appends it as a special token, then autoregressively generates trajectory tokens. This preserves full compatibility with the unified VA architecture while enabling effective long-horizon planning under strict token budgets.


\section{MEMORY-DEPENDENT DRIVING SCENARIOS AND EVALUATION FRAMEWORK}
\label{sec:scenarios}

We focus on scenarios where extended historical context plays a particularly important role in determining correct behavior. Prior studies~\cite{al2024autonomous} identify that critical driving decisions rely on behavioral cues captured within a 5-10 second temporal window, constituting extended context relative to standard driving policies that typically process only 1-2 seconds, and distinct from navigation tasks requiring long-term spatial memory over entire routes. We identify high-signal dynamic scenarios where extended context determines behavioral correctness and design metrics for these decision-critical outcomes beyond trajectory displacement.

\subsection{Stop-Controlled Intersections as Memory Testbeds}

\begin{figure}[tb]
  \begin{center}
    \centerline{\includegraphics[width=1.0\linewidth]{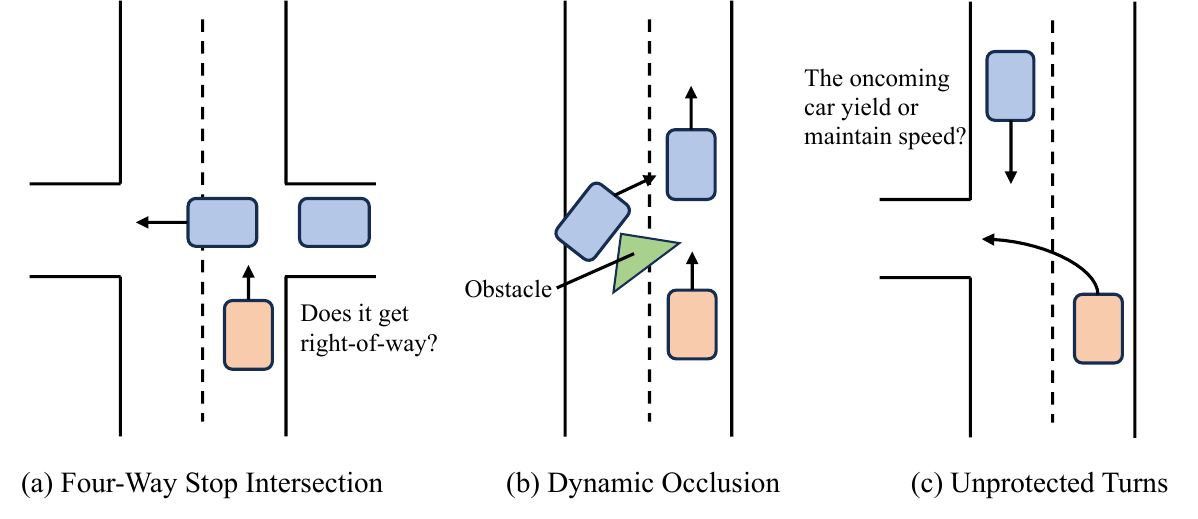}}
    \vspace{-2pt}
    \caption{\textbf{Illustration of high-signal dynamic scenarios.}}
    \label{fig:scenarios}
  \end{center}
  \vspace{-17pt}
\end{figure}

We identify three scenario classes (Fig.~\ref{fig:scenarios}) where extended historical context plays a prominent role: (1) all-way stops requiring right-of-way negotiation based on arrival order, (2) stop / yield signs requiring assessment of dynamic cross-traffic, and (3) unprotected turns requiring gap acceptance decisions. 
These share a fundamental characteristic that correct behavior depends on \textit{discrete decision correctness}, \ie, whether the vehicle stops when required and proceeds when appropriate, instead of trajectory smoothness.

\begin{itemize}[leftmargin=*]
\item \textbf{All-Way Stop Intersections.}
Multiple vehicles arrive at an intersection with all-way signs on all approaches. Right-of-way follows arrival order~\cite{al2024autonomous}, requiring the model to track which vehicles arrived first over 5-10 seconds to correctly infer yielding order and proceed promptly when gaining right-of-way.

\item \textbf{Dynamic Occlusion.}
Previously visible participants may become occluded as the ego approaches. The model must persist their state from seconds earlier rather than relying solely on current observation to avoid incorrectly assessing an occluded intersection as clear.

\item \textbf{Unprotected Turns.}
The ego turns across oncoming traffic without traffic light. The approaching vehicle observed seconds ago may decelerate to yield or maintain speed, and the ego must track its trajectory over several seconds to correctly decide whether to proceed or wait for a safe gap.

\end{itemize}

These three scenario classes encompass the core decision-making challenges in driving. According to~\cite{al2024autonomous}, fundamental skills tested here, \ie, errors in gap acceptance, right-of-way negotiation, and stopping behavior, account for approximately 40\% of intersection crashes.

\vspace{-3pt}\subsection{Behavioral Evaluation Metrics}

Traditional metrics like minADE~\cite{minade} are misaligned with these decision-critical scenarios. A rolling stop may achieve low minADE while constituting a safety violation~\cite{zhou2023matters}, whereas stopping correctly but slightly late incurs higher minADE despite being safer. We propose behavioral metrics directly assessing decision correctness across all three scenario classes:

\begin{itemize}[leftmargin=*]

\item \textbf{Stop Success Rate (Stop SR)} measures whether the vehicle achieves complete stop (velocity $< v_{\text{stop}}$) when required. For sustained stopping, we check whether the predicted trajectory maintains stopped state during ground-truth stopped periods within a temporal tolerance window.

\item \textbf{Go Success Rate (Go SR)} evaluates whether the vehicle proceeds after stopping rather than remaining indefinitely stopped. This includes assessing prompt departure after gaining right-of-way, ensuring the model does not create traffic flow disruptions.

\item \textbf{Roll-Through Rate} quantifies the percentage where the vehicle fails to achieve complete stop, instead performing a rolling stop (minimum velocity $\geq v_{\text{stop}}$). Rolling through is both illegal and dangerous~\cite{zhou2023matters,al2024autonomous}.

\item \textbf{Stop Position Error} measures spatial deviation between stop location and the designated stop line.

\item \textbf{Stop Duration Error} quantifies deviation between predicted and ground-truth stop duration from human driving.
\end{itemize}

\begin{table*}[t]
  \centering
  \caption{\textbf{Overall performance on stop-controlled intersection scenarios.} Under comparable token budgets, \alias achieves consistent improvements across all behavioral metrics while enabling 3.3$\times$ inference speedup and 2.7$\times$ memory reduction compared to uncompressed long-context processing (see Tab.~\ref{tab:efficiency}).}
  \vspace{-3pt}
  \resizebox{0.99\linewidth}{!}{
  \begin{tabular}{c|c|c|c|cc|cc}
    \toprule
    \textbf{Method} & \textbf{Settings} & \textbf{\makecell{Obs\\Tokens}} & \textbf{\makecell{Go SR ($\uparrow$)}}& \textbf{\makecell{Stop SR ($\uparrow$)}} & \textbf{\makecell{Roll-through\\Rate ($\downarrow$)}} & \textbf{\makecell{Stop Pos.\\Error ($\downarrow$)}} & \textbf{\makecell{Stop Dur.\\Err (sec) ($\downarrow$)}}\\
    \midrule
    \textbf{Standard Alpamayo~\cite{wang2025alpamayo}} & 1s 8 imgs & 1280  & 63.8\% \scriptsize{\raisebox{1pt}{$\pm 0.1\%$}} & 86.8\% \scriptsize{\raisebox{1pt}{$\pm 0.2\%$}} & 9.0\% \scriptsize{\raisebox{1pt}{$\pm 0.1\%$}} & 1.21 \scriptsize{\raisebox{1pt}{$\pm 0.02$}} & 0.50 \scriptsize{\raisebox{1pt}{$\pm 0.01$}} \\


    \textbf{Sparse Obs w/ Long Hist.} & 5s 8 imgs & 1280  & 62.0\% \scriptsize{\raisebox{1pt}{$\pm 0.1\%$}} & 86.2\% \scriptsize{\raisebox{1pt}{$\pm 0.1\%$}} & 9.3\% \scriptsize{\raisebox{1pt}{$\pm 0.0\%$}} & 1.25 \scriptsize{\raisebox{1pt}{$\pm 0.00$}} & 0.53 \scriptsize{\raisebox{1pt}{$\pm 0.00$}} \\
    
    \textbf{Dense Obs w/ Long Hist.} & 5s 40 imgs & 6400  & 61.9\% \scriptsize{\raisebox{1pt}{$\pm 0.1\%$}} & 85.8\% \scriptsize{\raisebox{1pt}{$\pm 0.2\%$}} & 9.9\% \scriptsize{\raisebox{1pt}{$\pm 0.1\%$}} & 1.26 \scriptsize{\raisebox{1pt}{$\pm 0.01$}} & 0.54 \scriptsize{\raisebox{1pt}{$\pm 0.00$}} \\

    \midrule
    
    \textbf{Compression w/o plan-align} & 5s 40 imgs & 1424  & 65.6\% \scriptsize{\raisebox{1pt}{$\pm 0.2\%$}} & 87.5\% \scriptsize{\raisebox{1pt}{$\pm 0.2\%$}} & 8.5\% \scriptsize{\raisebox{1pt}{$\pm 0.1\%$}} & 1.15 \scriptsize{\raisebox{1pt}{$\pm 0.01$}} & 0.49 \scriptsize{\raisebox{1pt}{$\pm 0.01$}} \\

    \midrule

    \bestcell{\textbf{\alias (Disc.)}} & \bestcell{5s 40 imgs} & \bestcell{1424} & \bestcell{68.2\% \scriptsize{\raisebox{1pt}{$\pm 0.3\%$}}} & \bestcell{\textbf{89.2\%} \scriptsize{\raisebox{1pt}{$\pm 0.1\%$}}} & \bestcell{\textbf{7.0\%} \scriptsize{\raisebox{1pt}{$\pm 0.2\%$}}} & \bestcell{1.15 \scriptsize{\raisebox{1pt}{$\pm 0.03$}}} & \bestcell{\textbf{0.48} \scriptsize{\raisebox{1pt}{$\pm 0.01$}}} \\
    
    \bestcell{\textbf{\alias (Cont.)}} & \bestcell{5s 40 imgs} & \bestcell{1424} & \bestcell{\textbf{68.3\%} \scriptsize{\raisebox{1pt}{$\pm 0.2\%$}}} & \bestcell{88.5\%} \scriptsize{\raisebox{1pt}{$\pm 0.1\%$}} & \bestcell{7.1\% \scriptsize{\raisebox{1pt}{$\pm 0.3\%$}}} & 
    \bestcell{\textbf{1.10} \scriptsize{\raisebox{1pt}{$\pm 0.02$}}} & 
    \bestcell{\textbf{0.48} \scriptsize{\raisebox{1pt}{$\pm 0.01$}}}
    \\
    \bottomrule
  \end{tabular}}
  \label{tab:overall}
  \vspace{-10pt}
\end{table*}

\section{EXPERIMENTS}

\subsection{Evaluation Settings}
\textbf{Dataset and Scenarios.}
We evaluate on the Alpamayo physical AI dataset~\cite{nvidia2025physicalai_autonomous_vehicles} with two complementary protocols: \textit{open-loop} evaluation on curated memory-dependent scenarios, and \textit{closed-loop} evaluation on general driving (Sec.~\ref{sec:closedloop}). We curate a subset from the dataset where ground-truth trajectories exhibit 
(1) deceleration below 1 m/s within 6.4 s, 
(2) a stopped state ($< 0.5$ m/s) for at least 0.5 s, and 
(3) subsequent acceleration indicating successful gap acceptance. 
This yields approximately 16\% of the dataset where behavioral correctness is unambiguous and memory-dependent reasoning essential. 
For open-loop evaluation, we extract a validation set of 20,000 clips (20s each, 200 frames at 10Hz, critical decision point at frame 50) and use the remainder for training.

\vspace{2pt}\textbf{Implementation Details.}
\label{sec:impl_details}
For all experiments, we use $T=20$ timesteps (5s at 4Hz), $N_{\text{cam}}=2$ cameras, and $N_{\text{img}}=160$ tokens per image. Without compression, this yields $N_{\text{raw}}=6{,}400$ vision tokens. Our hierarchical compression employs $K=3$ layers: Layer 1 ($n_1=4$ frames, $r_1=1$), Layer 2 ($n_2=5$ frames, $r_2=16$), and Layer 3 ($n_3=11$ frames, $r_3=80$), compressing to $N_{\text{compressed}}=1{,}424$ tokens ($4.5\times$ reduction). The driving latent dimension is $d_z=32$. For the posterior encoder, trajectories are quantized into 128 discrete tokens (2 per waypoint for 64 waypoints), then compressed to $N_{\text{agg}}=5$ tokens ($N_{\text{local}}=4$ local + 1 global). The VQ codebook size is $K=20$. For behavioral metrics, the stop velocity threshold is $v_{\text{stop}}=0.5$ m/s.

\vspace{2pt}\textbf{Baseline Settings.}
We compare \alias against baselines under varying history and token budgets: \textbf{Standard Alpamayo} retains only the recent 1s observation (8 frames, 1280 tokens), representing basic setting~\cite{wang2025alpamayo}; \textbf{Sparse Obs w/ Long Hist} extends context to 5s through sparse sampling (8 frames, 1280 tokens); \textbf{Dense Obs w/ Long Hist} maintains full 4Hz sampling over 5s without compression (40 frames, 6400 tokens); \textbf{Compression w/o plan-align} applies hierarchical compression without planning-aligned module (40 frames, 1424 tokens); and our \textbf{\alias (Disc./Cont.)} add planning-aligned variational compression using discrete FSQ-based or continuous latent encodings (1424 tokens). All methods are trained end-to-end on the same dataset.

\begin{table}[t]
\vspace{3pt}
\centering
\caption{\textbf{Closed-loop evaluation in Alpasim.}}
\vspace{-3pt}
\label{tab:alpasim}
\small
\begin{tabular}{l|cc}
\toprule
\multirow{2}{*}{\textbf{Metrics}}                                  & \multicolumn{1}{c}{\textbf{Baseline}} & \multicolumn{1}{c}{\textbf{Ours}} \\
 & 2s 18 imgs & 5s 40 imgs \\
\midrule
avg\_dist\_between\_incidents\_at\_fault & 0.49                                & 0.48                     \\
collision\_at\_fault                     & 0.04                                & 0.05                     \\
dist\_traveled\_m                        & 155.98                              & 152                      \\
min\_ade@0.5s(gt)                        & 3.77                                & 3.82                     \\
min\_ade@1.0s(gt)                        & 3.87                                & 3.91                     \\
min\_ade@2.5s(gt)                        & 4.14                                & 4.16                     \\
offroad                                  & 0.28                                & 0.27                     \\
offroad\_or\_collision\_at\_fault        & 0.32                                & 0.32                     \\
plan\_deviation                          & 0.79                                & 0.75                     \\
progress                                 & 0.73                                & 0.71                     \\
wrong\_lane                              & 0.23                                & 0.23                    \\
\bottomrule
\end{tabular}
\end{table}

\vspace{4pt}
\begin{table}[t]
\centering
\small
\caption{\textbf{Efficiency performance in Alpasim.}}
\label{tab:efficiency}
\vspace{-3pt}
\resizebox{1.0\linewidth}{!}{
\begin{tabular}{l|ccc}
\toprule
\multirow{2}{*}{\textbf{Efficiency Metrics}} & \multicolumn{1}{c}{\textbf{Baseline}} &  \textbf{Baseline (long)} & \multicolumn{1}{c}{\textbf{Ours}} \\
 & 2s 18 imgs & 5s 40 imgs & 5s 40 imgs \\
\midrule
Mean Time & 498.50 ms    & 1253.52 ms & 377.08 ms \\
Std  & 5.59 ms      &  27.99 ms  & 12.85 ms  \\
Median Time & 498.08 ms  & 1242.11 ms & 374.05 ms \\
Peak GPU Memory  & 5.94 GB  & 10.51 GB  & 3.95 GB\\
\bottomrule
\end{tabular}}
\vspace{-3pt}
\end{table}

\begin{table*}[t]
  \centering
  \caption{\textbf{Ablation on architecture.} All variants use 5s 40imgs with compression ratios 1-16-80. \rebuttal{Na\"{i}ve compression applies Q-former to visual tokens only; Compression w/o plan-aligned additionally conditions on past states and actions; COMPACT-VA further adds planning-aligned variational coupling, achieving optimal performance.}}
  \vspace{-2pt}
  \tabcolsep3.5pt
  \resizebox{0.95\linewidth}{!}{
  \begin{tabular}{c|ccc|cc|cc}
    \toprule
    \textbf{\makecell{Architecture\\Settings}} & \textbf{\makecell{Compression\\Module}} & \textbf{\makecell{History\\Info}} & \textbf{\makecell{Future\\Info}} & \textbf{\makecell{Go SR ($\uparrow$)}}& \textbf{\makecell{Stop SR ($\uparrow$)}} & \textbf{\makecell{Stop Pos.\\Error ($\downarrow$)}} & \textbf{\makecell{Stop Dur.\\Err (sec) ($\downarrow$)}} \\
    \midrule

     \textbf{No compression} & $\times$ & $\times$ & $\times$  & 61.9\% \scriptsize{\raisebox{1pt}{$\pm 0.1\%$}} & 85.8\% \scriptsize{\raisebox{1pt}{$\pm 0.2\%$}} & 1.26 \scriptsize{\raisebox{1pt}{$\pm 0.01$}} & 0.54 \scriptsize{\raisebox{1pt}{$\pm 0.00$}} \\

     \textbf{Na\"ive compression} & \checkmark & $\times$ & $\times$ & 63.5\% \scriptsize{\raisebox{1pt}{$\pm 0.2\%$}} & 86.6\% \scriptsize{\raisebox{1pt}{$\pm 0.2\%$}} & 1.21 \scriptsize{\raisebox{1pt}{$\pm 0.01$}} & 0.51 \scriptsize{\raisebox{1pt}{$\pm 0.01$}} \\

     \textbf{Compression w/o plan-aligned} & \checkmark & \checkmark & $\times$ & 65.6\% \scriptsize{\raisebox{1pt}{$\pm 0.2\%$}} & 87.5\% \scriptsize{\raisebox{1pt}{$\pm 0.2\%$}} & 1.15 \scriptsize{\raisebox{1pt}{$\pm 0.01$}} & 0.49 \scriptsize{\raisebox{1pt}{$\pm 0.01$}} \\
    
     \bestcell{\textbf{\alias (Ours)}} & \bestcell{\checkmark} & \bestcell{\checkmark} & \bestcell{\checkmark} & \bestcell{\textbf{68.3\%} \scriptsize{\raisebox{1pt}{$\pm 2.1\%$}}} & \bestcell{\textbf{88.5\%}} \scriptsize{\raisebox{1pt}{$\pm 0.1\%$}} & 
    \bestcell{\textbf{1.10} \scriptsize{\raisebox{1pt}{$\pm 0.02$}}} & 
    \bestcell{\textbf{0.48} \scriptsize{\raisebox{1pt}{$\pm 0.01$}}}
    \\
    \bottomrule
  \end{tabular}
  }
  \label{tab:arch_ablation}
  \vspace{-8pt}
\end{table*}


\begin{table}[t]
  \centering
  \caption{\rebuttal{\textbf{Ablation on hierarchical compression rates.} Each layer is described by the number of frames $n_k$ and cumulative compression ratio $r_k$.}}
  \vspace{-4pt}
  \begin{threeparttable}
  \resizebox{0.96\linewidth}{!}{
  \begin{tabular}{cc|cc|cc|c|cc}
    \toprule
    \multicolumn{2}{c|}{\textbf{Layer 1}} & \multicolumn{2}{c|}{\textbf{Layer 2}} & \multicolumn{2}{c|}{\textbf{Layer 3}} & \multirow{2}{*}{\textbf{\makecell{Token\\Num}}} & \multirow{2}{*}{\textbf{Go SR}} & \multirow{2}{*}{\textbf{Stop SR}} \\
    $n_1$ & $r_1$ & $n_2$ & $r_2$ & $n_3$ & $r_3$ & & & \\
    \midrule
    8 & 1 & ---& ---& ---& ---& 1280 & 63.8\% & 86.8\% \\
    \midrule
    8 & 2 & 10 & 16 & 22 & 80 & 784  & 64.6\% & 86.9\% \\
    4 & 1 & 10 & 16 & 26 & 80 & 792  & 66.1\% & 88.3\% \\
    \midrule
    8 & 1 & 10 & 16 & 22 & 80 & 1424 & \textbf{68.3\%} & 88.5\% \\
    8 & 1 & 10 &  8 & 22 & 80 & 1524 & 67.5\% & \textbf{88.9\%} \\
    8 & 1 & 20 & 16 & 12 & 80 & 1504 & 66.5\% & 88.2\% \\
    8 & 1 & 20 &  8 & 12 & 80 & 1704 & 67.1\% & 88.3\% \\
    \bottomrule
  \end{tabular}}
  \end{threeparttable}
  \label{tab:compression_rates}
\end{table}

\vspace{3pt}
\begin{table}[t]
  \centering
  \caption{\textbf{Ablation on history length.} Longer history improves performance, with 5s 40imgs peaking at go success rates, due to the base model's pretraining distribution.}
  \vspace{-3pt}
  \resizebox{0.92\linewidth}{!}{
  \begin{tabular}{c|c|cc|c}
    \toprule
    \textbf{\makecell{History Length}} & \textbf{\makecell{Token\\Num}} & \textbf{\makecell{Go SR}}& \textbf{\makecell{Stop SR}} & \textbf{Overall}\\
    \midrule
    \textbf{5s 20 imgs} & 712 & 65.6\%  & 88.2\% & 76.9\% \\
    \textbf{5s 40 imgs} & 1424 & \textbf{68.3\%} & 88.5\% & 78.2\% \\
    \textbf{5s 60 imgs} & 2136 & 66.6\% & 88.7\% & 77.7\% \\
    \textbf{5s 80 imgs} & 2848 & 68.2\% & \textbf{89.0\%} & \textbf{78.6\%} \\
    
    \bottomrule
  \end{tabular}}
  \label{tab:history_length}
  \vspace{-3pt}
\end{table}

\subsection{Overall Performances}
\label{sec:overall}

Table~\ref{tab:overall} presents the overall performance across all evaluated methods. We prioritize \textit{Go Success Rate} as the primary metric, as it directly tests whether the model maintains effective memory to determine when to proceed, requiring long-horizon memory to assess cross-traffic patterns and right-of-way. In contrast, \textit{Stop Success Rate}, while important, can often be achieved reactively by observing immediate deceleration trends without extended memory.

The standard Alpamayo already achieves competitive performance (63.8\% Go SR), demonstrating the effectiveness of its architecture.
Sparse observation with long history (62.0\%) underperforms, revealing that overly sparse sampling discards critical intermediate frames and actually hurts performance. 
Surprisingly, dense observation with full history (61.9\%) performs worst despite accessing all 40 frames, \rebuttal{as the quadratic attention cost over 6,400 raw vision tokens makes it difficult for the transformer to identify decision-critical cues, and implicit sequence-length assumptions inherited from the short-context Alpamayo checkpoint are only partially resolved by post fine-tuning.} Learned hierarchical compression without planning alignment (65.6\%) improves over Alpamayo, validating structured compression \rebuttal{avoids such token redundancy.}

Under comparable token budgets to standard baseline (1424 vs 1280 tokens), \alias achieves sizable improvements of 68.3\% Go SR (+4.5\% absolute over Alpamayo, +2.7\% over compression w/o plan-align), 22\% relative reduction in Roll-through Rate (7.0\% vs.~9.0\%), higher Stop SR (89.2\% vs.~86.8\%), and lower Stop Duration Error (0.48s vs.~0.50s). These gains are achieved while maintaining the real-time constraint—crucially, our approach also delivers substantial computational efficiency over long-context alternatives, \ie, 3.3$\times$ inference speedup and 2.7$\times$ memory reduction compared to uncompressed 5s 40imgs processing (Tab.~\ref{tab:efficiency}). This dual advantage (improved decision-making with enhanced efficiency) demonstrates that coupling compression with planning through cVAE framework enables retention of critical history context while remaining computationally practical for deployment.

\vspace{-6pt}\subsection{Closed-loop Evaluation}
\label{sec:closedloop}

While our open-loop evaluation stress-tests memory-critical stop-controlled scenarios, closed-loop evaluation validates general driving competence. We conduct closed-loop simulation on 910 diverse scenarios from the Physical AI AV NuRec dataset\footnote{\href{https://huggingface.co/datasets/nvidia/PhysicalAI-Autonomous-Vehicles-NuRec}{https://huggingface.co:nvidia/PhysicalAI-Autonomous-Vehicles-NuRec}} in Alpasim~\cite{alpasim2025}, a state-of-the-art neural-reconstruction based end-to-end simulator. We focus closed-loop evaluation on nominal driving scenarios, as there is currently a lack of sufficient reconstructions for stop-controlled intersections in the Alpasim environments.

\begin{figure}[t]
  \begin{center}
    \centerline{\includegraphics[width=0.96\linewidth]{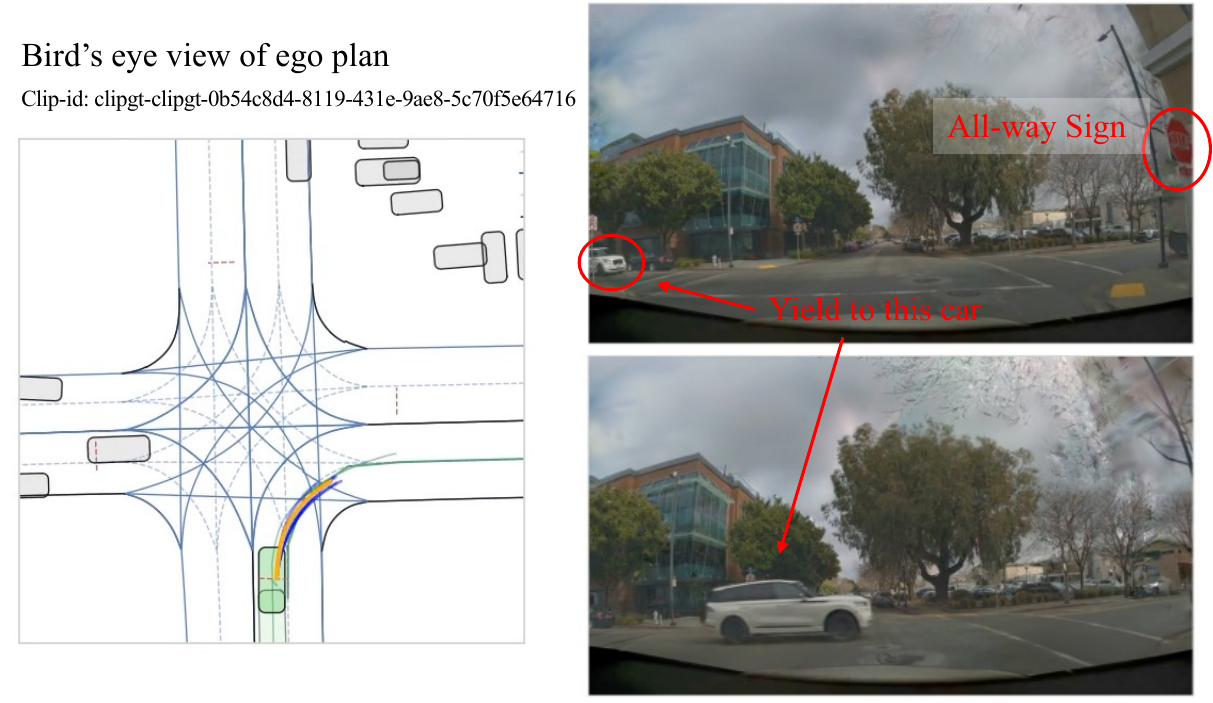}}
    \vspace{-2pt}
    \caption{\textbf{Closed-loop example at an all-way sign controlled right turn.} Our method correctly yields to the oncoming vehicle that arrived first. The bird's-eye view (left) shows both reference (yellow) and predicted (blue) trajectories stopping short, confirming the correct stop-and-wait decision. Camera views (right) capture the ego vehicle decelerating as the cross-traffic vehicle proceeds, demonstrating proper right-of-way negotiation enabled by maintaining historical context.}
    \label{fig:closed-loop}
  \end{center}
  \vspace{-13pt}
\end{figure}

We compare against the baseline (2s 18imgs), both trained on the same general driving data. Table~\ref{tab:alpasim} shows our method (5s 40imgs with compression) performs on par with the baseline across key safety metrics including, collision at fault, and average distance between incidents. Importantly, while maintaining this comparable driving performance with minimal token overhead, our approach delivers substantial computational efficiency gains.
We measure inference time and memory usage (averaged over 20 runs) on NVIDIA A100. Compared to the short-context baseline, our method achieves 1.32$\times$ faster inference (377ms vs 499ms) with 33\% lower peak GPU memory (3.95GB vs 5.94GB). The efficiency gains become more pronounced against the long-context without compression baseline processing the same temporal extent. Our compression delivers 3.3$\times$ speedup and 2.7$\times$ memory reduction. These results confirm \alias maintains general driving competence while improving memory-dependent decision-making (Sec.~\ref{sec:overall}) with reasonable computational overhead. A qualitative closed-loop example on all-way sign scenario is shown in Fig.~\ref{fig:closed-loop}.

\begin{table}[t]
\centering
\caption{\rebuttal{\textbf{\rebuttal{Ablation on intent token intervention.}} All variants share the same planning-aligned checkpoint. Only $z_\text{skill}$ is modified at inference time.}}
\vspace{-4pt}
\label{tab:intent_ablation}
\tabcolsep 2pt
\resizebox{1.0\columnwidth}{!}{%
\begin{tabular}{lccccc}
\toprule
\textbf{Intervention} & \textbf{Go SR} $\uparrow$ & \textbf{Stop SR} $\uparrow$ & \makecell{\textbf{Roll-through}\\(\%) $\downarrow$} & \textbf{\makecell{Stop Pos.\\Err $\downarrow$}} & \textbf{\makecell{Stop Dur.\\Err (s) $\downarrow$}} \\
\midrule
None (ours)  & \textbf{68.2\%} & \textbf{89.2\%} & \textbf{7.0} & \textbf{1.151} & \textbf{0.477} \\
Mean         & 64.7\% & 87.5\% & 8.2 & 1.288 & 0.493 \\
Shuffle      & 66.1\% & 88.4\% & 7.6 & 1.214 & 0.500 \\
Randomize    & 63.7\% & 88.0\% & 8.2 & 1.254 & 0.498 \\
Remove       & 60.3\% & 86.3\% & 9.2 & 1.261 & 0.512 \\
\bottomrule
\end{tabular}%
}
\end{table}

\vspace{-8pt}\subsection{Ablation Study and Analysis}
\label{sec:ablation}

\textbf{Ablation on Architecture.}
Table~\ref{tab:arch_ablation} validates each component using 5s 40imgs with compression ratios 1-16-80 (8 frames at 160 tokens, 10 at 10 tokens, 22 at 2 tokens). Without compression, Go SR is only 61.9\% despite all 40 frames, confirming raw token abundance hinders performance. Na\"ive compression improves to 63.5\%, \rebuttal{where the Q-former compresses visual tokens without any additional conditioning signal.} Standard compression incorporating historical trajectory as Q-former conditioning achieves 65.6\%, \rebuttal{where the ego's past states and actions are encoded into a single token concatenated with visual tokens as an extra conditioning input,} demonstrating trajectory-conditioned compression retains decision-relevant features. 
Adding planning-aligned variational coupling, COMPACT-VA reaches 68.3\% by appending the predicted future latent as a special token through the conditional VQ-VAE framework, \rebuttal{which couples compression quality with trajectory prediction and enables the model to retain planning-relevant history.}


\textbf{Ablation on Learned Skills \rebuttal{and Intent Token Usage.}}
To validate effective skill learning, we track VQ codebook utilization ($K=20$) via an exponentially weighted moving average (EWMA) of per-step skill probabilities:
\vspace{-2pt}
\begin{equation}
    p_t^{\text{smooth}}(k) = \alpha \cdot p_{t-1}^{\text{smooth}}(k) + (1-\alpha) \cdot p_t^{\text{batch}}(k),
\end{equation}
where $\alpha = 0.99$ and a skill $k$ is active if $p_t^{\text{smooth}}(k) > 1/K$. The model consistently activates 15-17 out of 20 skills across four seeds, achieving 80\% codebook utilization and 
without mode collapse. 
\rebuttal{To verify the policy actively uses the intent token, we apply four inference-time interventions (mean, shuffle, randomize, remove) to $z_\text{skill}$ on top of the planning-aligned checkpoint (Table~\ref{tab:intent_ablation}). All interventions consistently degrade performance across all metrics, with Go~SR dropping from 68.2\% to as low as 60.3\%, confirming that the policy actively exploits the semantic information carried by the intent token rather than bypassing it. The \emph{remove} variant even falls below the \emph{Compression w/o plan-aligned} baseline (65.6\%), as backbone co-adaptation during training makes inference-time removal more disruptive than training without it.}


\textbf{Ablation on Different Compression Rates.}
Table~\ref{tab:compression_rates} examines hierarchical compression configurations under 5s 40imgs, with each layer described by $n_k$ and $r_k$. \rebuttal{Several design principles emerge to guide the selection of $r_k$. The most recent layer should retain full token resolution ($r_1 = 1$), as compressing layer 1 to $r_1 = 2$ consistently degrades Go SR from 68.3\% to 64.6\%.} Allocating more images to layer 1 is beneficial where $(8{\times}160)$ (68.3\%) outperforms $(4{\times}160)$ (66.1\%), \rebuttal{and the token counts follow a roughly three-order-of-magnitude progression (160\,/\,10\,/\,2 per image) from recent to distant frames.} Increasing layer 2 from 10 to 20 frames degrades Go SR under fixed token budgets, suggesting over-extending distant history with sparse allocation fails to capture useful information. 
%
\rebuttal{Importantly, every compressed variant outperforms the no-compression baseline (61.9\% Go SR), confirming the framework is robust to precise ratio values and that the design principle of allocating more tokens to recent frames matters more than exact hyperparameter tuning.}

\vspace{2pt}
\textbf{Ablation on History Length.}
Table~\ref{tab:history_length} evaluates different frame counts with fixed compression ratios (1-16-80) and layer proportions (4:5:11). While 5s 40imgs (1424 tokens) achieves the highest Go SR (68.3\%), 5s 80imgs (2848 tokens) achieves competitive result (68.2\%) and the best overall score (78.6\%), showing that longer history generally benefits performance. The slight performance variation suggests that history length interacts with model capacity and pretraining distribution, \ie, our fine-tuning approach is constrained by the base model's exposure to specific history lengths during pretraining, affecting optimal length within explored range.

\section{CONCLUSION}
\label{sec:conclusion}

This work presents \alias, a planning-aligned working memory framework for monolithic autonomous driving policies. By coupling compression with trajectory prediction through conditional VQ-VAE, we address the fundamental limitation that unified VA/VLA policies either lack explicit memory mechanisms or rely on rule-based compression unable to guarantee retention of decision-relevant history.
%
Through systematic evaluation on memory-critical stop-controlled scenarios, we demonstrate that planning-aligned compression achieves sizable improvements in decision correctness while maintaining real-time token budgets. Our working memory approach with bounded context windows is well-aligned with driving, where critical decisions typically depend on behavioral cues within 5-10 seconds. We believe this work advances the field toward effective memory for VA/VLA policies, demonstrating task-aware working memory is key to reasoning in diverse scenarios.

\textbf{Future Work} could explore higher-complexity scenarios (\eg,~severe occlusions or multiple contenders), recurrent memory mechanisms and state-space models to extend planning aligned compression to broader physical AI domains.


\bibliographystyle{./IEEEtran} 
\bibliography{./IEEEabrv,./IEEEexample}

\end{document}